\documentclass[runningheads]{llncs}
\usepackage[T1]{fontenc}
\usepackage{graphicx}
\usepackage{amsmath,amssymb}
\usepackage{booktabs}
\usepackage[table]{xcolor}
\usepackage{multirow}
\usepackage{algorithm}
\usepackage{algorithmic}
\usepackage{url}
\usepackage{hyperref}
\usepackage{makecell}
\usepackage{placeins}

\begin{document}
\title{DINE: Distance Is Not Enough\\Learning Global Deformation Priors for Robust Soft-Tissue Point Cloud Registration}
\titlerunning{DINE: Distance Is Not Enough}

%
%
\author{
Sara Monji-Azad\inst{1}\thanks{Corresponding author: sara.monjiazad@medma.uni-heidelberg.de}
\and Rohit Beer\inst{1}
\and Marvin Kinz\inst{2,1}
\and Claudia Scherl\inst{3,4}
\and J\"urgen Hesser\inst{1,5,6,7}
}
\authorrunning{S. Monji-Azad et al.}

\institute{
Mannheim Institute for Intelligent Systems in Medicine (MIISM), Medical Faculty Mannheim, Heidelberg University, Mannheim, Germany
\and Department of Radiation Oncology, Brigham and Women's Hospital, Dana-Farber Cancer Institute, Harvard Medical School, Boston, Massachusetts, USA
\and Department of Otorhinolaryngology, Head and Neck Surgery, University Medical Center Mannheim, Medical Faculty Mannheim, Heidelberg University, Mannheim, Germany
\and Department of Otolaryngology, Head and Neck Surgery, Campus Klinikum Bielefeld Mitte, University Hospital OWL of Bielefeld University, Bielefeld, Germany
\and Interdisciplinary Center for Scientific Computing (IWR), Heidelberg University, Heidelberg, Germany
\and Central Institute for Computer Engineering (ZITI), Heidelberg University, Heidelberg, Germany
\and CZS Heidelberg Center for Model-Based AI, Heidelberg University, Mannheim, Germany
}

\maketitle              

\begin{abstract}
Non-rigid point cloud registration is central to soft-tissue shape analysis, but large deformations, noise, and outliers make correspondence estimation challenging. Most learning-based methods rely on local objectives such as Chamfer distance, which encourage point-wise proximity but do not constrain the global plausibility of the predicted deformation field. We address this limitation with DINE, a maximum a posteriori framework that augments distance-based registration with a learned statistical prior over displacement vector fields. DINE is applied to two registration backbones, Robust-DefReg and DefTransNet, using a two-stage strategy: a first-stage model is trained with Chamfer distance, its predicted deformation fields are used to estimate a prior, and the model is then refined with a combined distance and negative log-prior objective. We compare a full-field PCA Gaussian prior with a per-vector normalizing-flow prior. Experiments on DeformedTissue and SynBench show lower mean Chamfer distance under deformation and corruption. On DeformedTissue, DINE-PCA reduces Chamfer distance by approximately 27--69\% relative to the corresponding Stage-1 backbone across deformation levels, and improves robustness by up to 66\% for outliers and 83\% for Gaussian noise. On SynBench, improvements are modest at the smallest deformation levels and reach approximately 59--79\% from moderate to severe deformation. These results suggest that global deformation plausibility is an important constraint for reliable soft-tissue point cloud registration. (The code will be published soon.)


\end{abstract}

\section{Introduction}
Three-dimensional shape representations are increasingly important in medical image computing, surgical planning, image-guided intervention, and digital-twin systems~\cite{carter2005softtissue,miga2016computational,mannle2023artificial,monji2024point}. In applications involving soft tissue, geometry is not static. Tissue can deform due to manipulation, gravity, swelling, resection, or temperature-induced change, and reliable non-rigid registration is required to align shapes acquired at different states~\cite{carter2005softtissue,miga2016computational,han2024organ,monji2024point}. Point clouds are a natural representation for such surfaces because they can be obtained from depth sensors, stereo reconstruction, photogrammetry, or processed meshes~\cite{monji2023review,weber2024arpc,monji2025deftransnet}. Yet registering deformable point clouds remains challenging when deformation is large, non-uniform, noisy, or only partially observed~\cite{monji2023review,monjiazed2024synbench,monji2024robust,monji2025deftransnet}.

Learning-based non-rigid registration methods often optimize a point-wise distance objective such as Chamfer distance~\cite{fan2017pointset,monji2024robust,monji2025deftransnet}. Such objectives are attractive because they do not require explicit point-to-point correspondences and can be evaluated directly on unordered point sets~\cite{fan2017pointset,myronenko2010point,monji2023review}. However, they primarily enforce local proximity between the registered source and the target. They do not specify whether the full deformation vector field (DVF) is anatomically, statistically, or physically plausible. Under large deformation, several displacement fields can produce a similar local distance, although only a subset corresponds to coherent tissue motion. Classical registration has long addressed this ambiguity using smoothness, elasticity, diffeomorphism, or statistical priors~\cite{bookstein1989principal,rueckert1999nonrigid,modersitzki2004numerical,fischer2008ill,myronenko2010point}. Modern neural point cloud registration can benefit from the same principle, but the prior should be compatible with learned DVFs and unordered geometric data~\cite{monji2023review,monji2024robust,monji2025deftransnet}.

In medical shape analysis, prior information is often expressed through statistical shape or deformation models. Classical statistical shape models use low-dimensional representations to capture valid anatomical variability~\cite{cootes1995active,heimann2009statistical}. Probabilistic and learning-based deformable registration methods introduce explicit regularizers, latent variables, or velocity-field constraints to model plausible transformations~\cite{leow2007statistical,dalca2019unsupervised,yang2017quicksilver,balakrishnan2019voxelmorph}. In point clouds, recent registration networks have strong representational capacity, but when trained only with local alignment losses they may still produce globally inconsistent displacement fields. This motivates an explicit bridge between learning-based point cloud registration and statistical deformation modeling.

We propose DINE, a prior-regularized learning framework for non-rigid point cloud registration with learned deformation priors. Starting from a maximum a posteriori (MAP) decomposition, the data term corresponds to the likelihood of the target given the deformed source, while the deformation prior penalizes unlikely DVFs. Our method uses a two-stage strategy. In the first stage, a registration backbone is trained using Chamfer distance. The predicted DVFs from this baseline define an empirical deformation distribution. These Stage-1 DVFs are not treated as ground-truth deformation fields; rather, they provide an estimate of dominant deformation patterns observed by a trained registration model. In the second stage, the same backbone is refined using the original Chamfer term plus a negative log-prior computed from this distribution, which suppresses statistically rare or irregular DVFs during refinement.

Our main contribution is not a new registration backbone, but a general prior-based training formulation that can be applied to existing non-rigid point cloud registration networks. We instantiate the formulation with two priors. The first is a PCA Gaussian prior over the full vectorized DVF, which captures correlations between point displacements and therefore models global deformation modes. The second is a RealNVP normalizing-flow prior over individual displacement vectors, which provides a more expressive local density but does not capture inter-point structure. By comparing both variants, we show that global spatial correlation is crucial for robust soft-tissue shape registration. The contributions of this work are:
\begin{enumerate}
    \item We formulate non-rigid point cloud registration with learned deformation priors as MAP estimation, making explicit the role of a negative log-prior in addition to local geometric alignment.
    \item We introduce a two-stage, architecture-agnostic regularization strategy based on learned DVF statistics, with a full-field PCA Gaussian prior as the main model and a per-vector flow prior as a controlled ablation.
    \item We evaluate the method on DeformedTissue, a real soft-tissue deformation dataset, and SynBench, a controlled synthetic benchmark dataset, showing that the PCA-based DINE variant consistently improves registration under large deformation, Gaussian noise, and outlier contamination.
\end{enumerate}

\section{Related Work}

\textbf{Classical and probabilistic registration.}
Classical registration methods such as ICP~\cite{besl1992method} and Coherent Point Drift (CPD)~\cite{myronenko2010point} align point sets by iteratively estimating correspondences and transformations. CPD is particularly relevant because it interprets registration probabilistically and regularizes the deformation field through a coherent motion model. In image registration, free-form deformations, thin-plate splines, elastic models, and diffeomorphic formulations have provided principled ways to constrain transformations~\cite{bookstein1989principal,rueckert1999nonrigid,modersitzki2004numerical,dalca2019unsupervised}. DINE follows this tradition but learns the deformation prior from predicted point-cloud DVFs rather than prescribing a fixed regularizer.

\textbf{Learning-based non-rigid point cloud registration.}
Deep point cloud registration replaces hand-crafted optimization components with neural features, correspondence modules, and deformation predictors. PointNetLK~\cite{aoki2019pointnetlk}, Deep Closest Point~\cite{wang2019deep}, Lepard~\cite{li2022lepard}, RegTR~\cite{yew2022regtr}, and Neural Deformation Pyramid~\cite{li2022non} illustrate the broad progress in rigid and deformable point cloud registration. Recent surveys further summarize non-rigid transformations and learning-based 3D point cloud registration~\cite{monji2023review}. In the soft-tissue setting, Robust-DefReg~\cite{monji2024robust} and DefTransNet~\cite{monji2025deftransnet} provide graph- and transformer-based backbones for non-rigid registration of deformable surfaces. We use these two backbones to test whether a learned prior improves existing architectures without modifying their internal design.

\textbf{Medical shape and deformation statistics.}
Statistical shape models have a long history in medical image analysis, where PCA and related low-dimensional models encode valid anatomical variability~\cite{cootes1995active,heimann2009statistical}. Deformation priors have also been used in large-deformation image registration and learned registration frameworks~\cite{leow2007statistical,yang2017quicksilver,balakrishnan2019voxelmorph,dalca2019unsupervised}. Our full-field PCA prior follows the same statistical modeling idea, but applies it to complete point-cloud DVFs. This allows the model to penalize globally implausible displacement configurations while remaining lightweight and easy to add to existing networks.

\textbf{Soft-tissue digital twins and DeformedTissue.}
Soft-tissue deformation is a central challenge in surgical digital twins, particularly for head and neck surgery where tissue shift can change the relationship between preoperative planning and intraoperative geometry. Experimental pig-head cadaver studies have used HoloLens and handheld 3D scanning to capture resection cavities and cut tissue pieces under controlled deformation~\cite{mannle2023artificial,monji2024point}. DefTransNet later used DeformedTissue as a real-world test set for non-rigid point cloud registration~\cite{monji2025deftransnet}. In this work, DeformedTissue is used as the primary biomedical dataset to evaluate whether statistical DVF priors improve soft-tissue registration.

\textbf{Robust geometric losses and learned priors.}
Chamfer distance is widely used for point cloud completion and registration, but it is sensitive to outliers, sampling density, and ambiguous nearest-neighbor matches~\cite {azad2025coarse}. Several variants improve robustness, including Density-aware Chamfer Distance~\cite{wu2021dcd}, Hyperbolic Chamfer Distance~\cite{lin2023hypercd}, weighted Chamfer losses~\cite{lin2024weightedcd}, and Flexible-weighted Chamfer Distance~\cite{li2025fcd}. These methods modify the metric itself. Our method is complementary: it keeps the geometric data term but adds a deformation prior that constrains the global solution space. Normalizing flows provide an alternative way to model learned densities~\cite{dinh2017density,papamakarios2021normalizing}; we therefore include a RealNVP prior as a comparison to the full-field PCA model.

\section{Method}

\subsection{Problem Formulation}

Let the source and target point clouds be $X=\{x_i \in \mathbb{R}^3\}_{i=1}^{N}$ and $Y=\{y_j \in \mathbb{R}^3\}_{j=1}^{M}$, respectively.
A non-rigid registration network $f_\theta$ predicts a deformation vector field
\begin{equation}
D = f_\theta(X,Y) = \{d_i \in \mathbb{R}^3\}_{i=1}^{N},
\end{equation}
which produces the registered source
\begin{equation}
\hat{X} = X + D = \{x_i + d_i\}_{i=1}^{N}.
\end{equation}
We also use the vectorized representation $D \in \mathbb{R}^{3N}$ by concatenating all displacement vectors. In our data representation, the source point order is preserved across samples. Therefore, the $i$-th displacement vector corresponds to the same source point index under the fixed sampling convention, making vectorized DVFs comparable across the training set and enabling full-field PCA to model meaningful inter-point correlations.

The standard distance-only objective is the symmetric Chamfer distance
\begin{equation}
\mathcal{L}_{\mathrm{CD}}(\hat{X},Y) =
\frac{1}{N}\sum_{i=1}^{N}\min_j \|\hat{x}_i-y_j\|_2^2
+
\frac{1}{M}\sum_{j=1}^{M}\min_i \|y_j-\hat{x}_i\|_2^2 .
\end{equation}
Although this objective enforces point-wise proximity, it does not distinguish between globally coherent and incoherent DVFs when both yield similar local distances.

\subsection{MAP View of Deformation-Regularized Registration}
We view registration as estimating a DVF $D$ given a source point cloud $X$ and
a target point cloud $Y$. By Bayes' rule,
\begin{equation}
 p(D \mid X,Y) \propto p(Y \mid X,D)p(D),
\end{equation}
where $p(Y \mid X,D)$ measures data fidelity after deforming $X$ by $D$, and $p(D)$ denotes a deformation prior. Here we use an unconditional prior over DVFs; more generally, this term could also be written as $p(D \mid X)$ if the prior is explicitly conditioned on the source geometry. The MAP estimate is therefore
\begin{equation}
\hat{D} = \arg\min_D
\underbrace{-\log p(Y \mid X,D)}_{\mathrm{data\ term}}
-
\underbrace{\log p(D)}_{\mathrm{log\mbox{-}prior\ term}} .
\label{eq:map}
\end{equation}

To obtain a practical point-cloud registration loss, we interpret the point coordinates in Euclidean space and assume isotropic squared residual penalties between matched points. Since explicit correspondences are unknown, the nearest-neighbor assignments used in the symmetric Chamfer distance provide a standard surrogate for this data-fidelity term. This should be understood as a modeling assumption rather than an exact probabilistic observation model.
Optimizing only the Chamfer distance therefore corresponds to omitting an explicit deformation-prior term from the objective. In this case, the loss penalizes local geometric mismatch but does not directly discourage globally unlikely or incoherent deformation fields. We therefore optimize
\begin{equation}
\mathcal{L} = \mathcal{L}_{\mathrm{CD}} + \lambda \mathcal{L}_{\mathrm{prior}},
\label{eq:loss}
\end{equation}
where $\lambda$ controls the strength of the learned deformation prior.

\subsection{Two-Stage Training Framework}

The proposed framework is summarized in Algorithm~\ref{alg:method}. Stage 1 trains a registration backbone using only the Chamfer distance. The predicted DVFs from the training set are then collected and used to estimate a statistical deformation prior. Stage 2 initializes the same backbone from the Stage-1 weights and fine-tunes it with the combined objective in Eq.~\eqref{eq:loss}. The prior is estimated only from Stage-1 DVFs on the training split and is then frozen during Stage-2 refinement; no test-set DVFs are used for prior estimation. This strategy is architecture-agnostic and can be applied to different registration networks.

\begin{figure}[t]
\centering
\includegraphics[width=\textwidth]{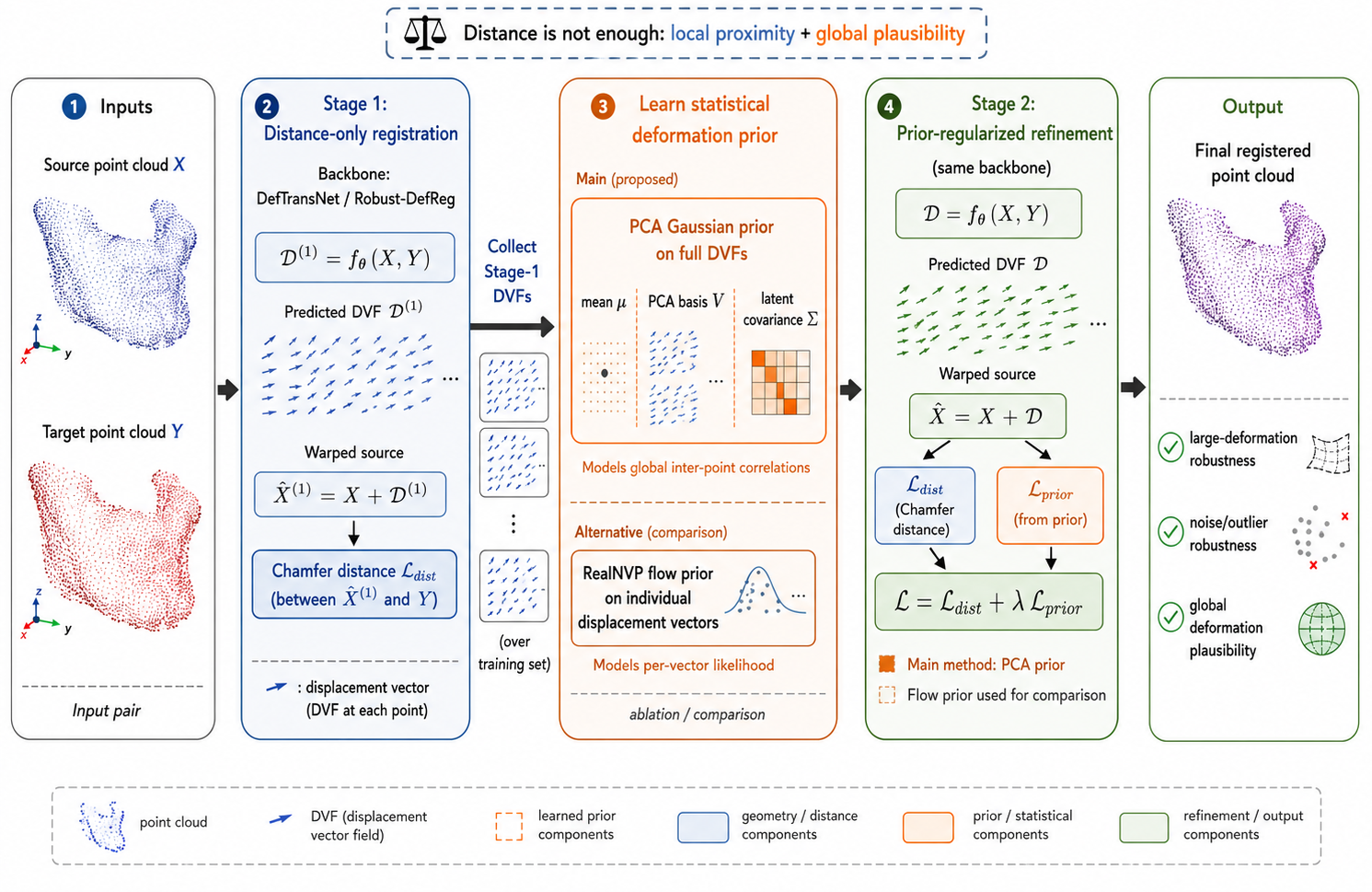}
\caption{Overview of DINE. A distance-only registration backbone is first trained to predict deformation vector fields (DVFs). The collected Stage-1 DVFs are used to learn a statistical deformation prior, with the PCA Gaussian prior over full DVFs as the main model and a per-vector RealNVP prior as a controlled ablation. In Stage 2, the same backbone is refined using a combined distance and prior-regularized objective, encouraging both local alignment and global deformation plausibility.}
\label{fig:method}
\end{figure}

\begin{algorithm}[t]
\caption{Prior-Regularized Non-rigid Point Cloud Registration}
\label{alg:method}
\begin{algorithmic}[1]
\STATE Train a registration network $f_\theta$ with $\mathcal{L}_{\mathrm{CD}}$.
\STATE For each training pair $(X^{(k)},Y^{(k)})$, collect $D^{(k)}=f_\theta(X^{(k)},Y^{(k)})$.
\STATE Estimate a deformation prior $p(D)$ from $\{D^{(k)}\}_{k=1}^{K}$.
\STATE Initialize Stage 2 from the Stage 1 network.
\STATE Refine the network using $\mathcal{L}=\mathcal{L}_{\mathrm{CD}}+\lambda[-\log p(D)]$.
\STATE Return the prior-regularized registration model.
\end{algorithmic}
\end{algorithm}

\subsection{Full-Field Principal Component Analysis (PCA) Gaussian Prior}
The PCA-based prior, one of the two priors considered in this work, operates on the complete vectorized DVF, obtained by concatenating all $N$ displacement vectors into a single vector in $\mathbb{R}^{3N}$. The Gaussian form of this prior is a modeling assumption in the PCA latent space, chosen to obtain a simple and differentiable statistical deformation prior. This construction assumes an index-consistent point representation. In our experiments, this consistency is obtained during preprocessing by using a fixed source sampling convention for each registration pair. PCA is applied to deformation vectors whose entries have a fixed source-index meaning, i.e., the same entry of the vectorized DVF corresponds to the same source-point coordinate across samples. Given Stage 1 deformation vector fields $\{D^{(k)}\}_{k=1}^{K}$, we compute the empirical mean $\mu$ and coordinate-wise standard deviation $\sigma$ across the training set, and normalize each field as
\begin{equation}
\tilde{D}^{(k)} = \frac{D^{(k)}-\mu}{\sigma+\epsilon_\sigma}.
\end{equation}
Here $\epsilon_\sigma$ is a small constant used only to avoid division by near-zero standard deviations. PCA is applied to the normalized fields, yielding a basis $V \in \mathbb{R}^{3N \times n_c}$ and latent coordinates
\begin{equation}
z^{(k)} = V^\top \tilde{D}^{(k)}.
\end{equation}
The covariance in latent space is estimated from the latent coordinate vectors $z^{(k)}$ across the training set and denoted by the regularized covariance matrix $\hat{\Sigma}_{\mathrm{reg}}$. Shrinkage and a small diagonal regularization term are applied during this covariance estimation to ensure numerical stability~\cite{ledoit2004well}.

For a Stage 2 prediction $D$, we compute $\tilde{D}=(D-\mu)/(\sigma+\epsilon_\sigma)$ and $z=V^\top\tilde{D}$. The prior penalty is the Mahalanobis distance
\begin{equation}
\mathcal{L}_{\mathrm{prior}}^{\mathrm{PCA}} =
z^\top \hat{\Sigma}_{\mathrm{reg}}^{-1} z .
\label{eq:pca_prior}
\end{equation}
This corresponds to the negative log-prior under a learned linear Gaussian deformation model, up to additive constants. Here, the Gaussian model should be understood as the assumed latent-space prior rather than as a claim that the true deformation distribution is exactly Gaussian. PCA on the collected DVFs can be interpreted as a maximum-likelihood fit of a low-dimensional Gaussian latent deformation model, making the prior a lightweight and differentiable statistical deformation model. Because the prior is defined over the full DVF, it captures spatial correlations between point displacements. This is important for tissue deformation, where neighboring regions usually move in correlated ways.

\subsection{Normalizing-Flow Prior}

As an alternative, we model individual displacement vectors $d_i \in \mathbb{R}^3$
using a RealNVP-style normalizing flow~\cite{dinh2017density}. Let $Q_\phi$ be
the learned density over individual displacement vectors. The
normalizing flow maps each displacement vector through invertible coupling layers
to a standard Gaussian distribution, giving
\begin{equation}
Q_\phi(d) = p_0(g_\phi(d))\left|\det \frac{\partial g_\phi}{\partial d}\right|.
\end{equation}
The normalizing flow is trained by maximum likelihood on pooled Stage 1 displacement vectors, which are treated as samples from the empirical Stage-1 displacement distribution. The learned invertible mapping $g_\phi$ maps this empirical distribution to a standard Gaussian distribution $\mathcal{N}(0,I)$. During Stage 2, the flow is frozen, meaning that its parameters are kept fixed and are not updated by gradients, and the prior term is
\begin{equation}
L^{\mathrm{flow}}_{\mathrm{prior}}
=
-\frac{1}{|\mathcal{V}|}
\sum_{i\in\mathcal{V}}
\log Q_{\phi}(d_i),
\end{equation}
where $\mathcal{V}$ denotes the set of valid point indices, i.e., the source points for which a displacement vector is used
in the prior computation. In the absence of masking, this corresponds to all source points. This prior can model a flexible distribution over local displacement vectors, but it cannot directly capture whether the full collection of vectors forms a spatially coherent deformation field, because the likelihood is evaluated independently for each displacement vector and does not model correlations between different point displacements.

\section{Experiments}

\subsection{Datasets and Implementation}

\begin{figure}[t]
\centering
\includegraphics[width=0.96\textwidth]{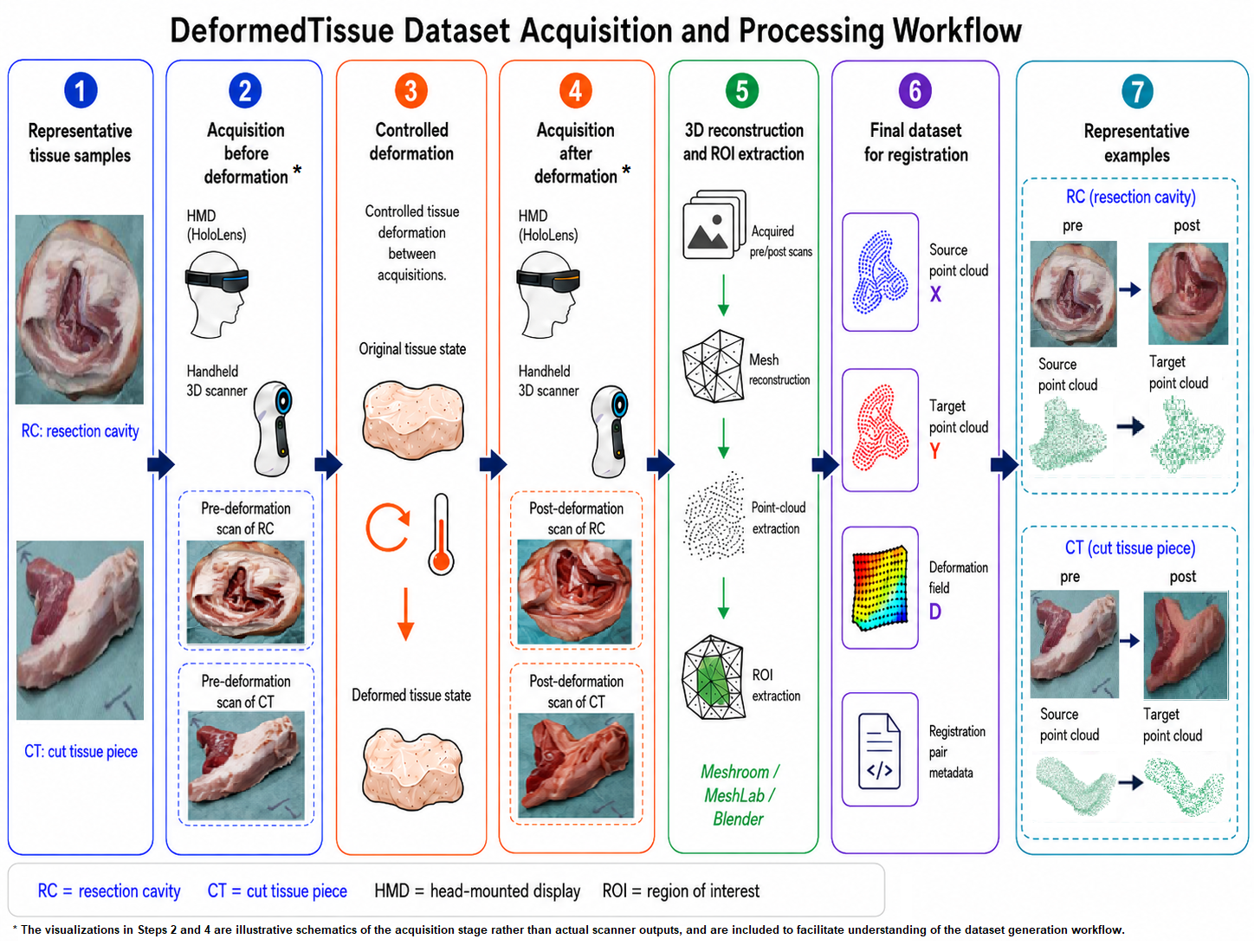}
\caption{DeformedTissue dataset acquisition and processing workflow. Representative resection cavities (RC) and cut tissue pieces (CT) are acquired before and after controlled deformation using head-mounted-display and handheld 3D scanning. The acquired data are reconstructed and processed into meshes, point clouds, regions of interest, and source-target registration pairs for non-rigid registration. The scan views in Steps~2 and~4 are schematic graphical illustrations added for conceptual clarity and are not actual raw scanner outputs. Full details of the dataset generation, scanning protocol, deformation procedure, and 3D processing pipeline are provided in previous DeformedTissue studies~\cite{monji2025deftransnet,mannle2023artificial,monji2024point}.}
\label{fig:dataset}
\end{figure}

\textbf{DeformedTissue.}
We use DeformedTissue as the primary biomedical evaluation dataset. It contains real soft-tissue shapes captured before and after controlled deformation induced by heating, using HoloLens and handheld 3D scanning in a pig-head cadaver study~\cite{mannle2023artificial,monji2024point}. The resulting 2.5D meshes are processed into source-target point cloud pairs. The dataset contains seven deformation levels, $0.1,0.2,\ldots,0.7$, with additional noisy and outlier-contaminated variants. It therefore provides a relevant testbed for non-rigid shape registration under real soft-tissue geometry, large deformation, and practical acquisition artifacts. Figure~\ref{fig:dataset} summarizes the acquisition and processing workflow. Full details of the dataset generation, scanning protocol, deformation procedure, and 3D processing pipeline are provided in previous DeformedTissue studies~\cite{monji2025deftransnet,mannle2023artificial,monji2024point}.

\textbf{SynBench.}
We further evaluate on SynBench, a controlled synthetic benchmark dataset for non-rigid point cloud registration~\cite{DataSynBench,monji2024robust}. It is connected to earlier simulation tools for soft-body and deformable point-cloud data generation~\cite{monji2023simtool}. SynBench contains source-target point cloud pairs generated from synthetic 3D shapes with deformation levels $\sigma \in \{0.1,0.2,\ldots,0.8\}$ produced by Gaussian radial basis functions. We use this dataset to isolate the effect of increasing deformation magnitude under controlled conditions and to verify whether the proposed prior generalizes beyond the soft-tissue dataset. All point clouds are resampled to 1024 points and the original train-test split is used.

\textbf{Backbones and training.}
We evaluate DINE on two learning-based registration backbones: Robust-DefReg~\cite{monji2024robust} and DefTransNet~\cite{monji2025deftransnet}. Stage 1 models are trained with Chamfer-distance supervision. Stage 2 models are initialized from Stage 1 and optimized with Eq.~\eqref{eq:loss}. The PCA prior uses $n_c=64$ components, Ledoit-Wolf shrinkage with $\alpha=0.05$, normalization constant $\epsilon_\sigma$ for standard-deviation scaling, and covariance diagonal regularization $\epsilon_\Sigma=10^{-3}$. The flow prior uses RealNVP coupling layers with hidden dimension 256, trained on collected Stage-1 displacement vectors. Both priors are estimated on training-set DVFs and kept fixed during Stage-2 optimization. The prior weight $\lambda$ is selected once by a validation search and then kept fixed for all datasets, backbones, and corruption settings, avoiding dataset-specific or backbone-specific tuning. Optimization uses AdamW~\cite{loshchilov2019decoupled}. All Chamfer distances are reported in units of \(10^{-4}\); lower is better.

\subsection{Results}

All values in the following tables are reported as mean $\pm$ standard deviation of Chamfer distance (CD); lower values indicate better registration accuracy.

\textbf{Increasing deformation.}
Tables~\ref{tab:deftissue_deformation} and~\ref{tab:synbench_deformation} report registration accuracy under increasing deformation magnitude. The methods are ordered by backbone, which allows a direct comparison between each Stage 1 baseline and its DINE-regularized variant.

On DeformedTissue, DINE-PCA obtains lower mean CD than the corresponding Stage-1 model for both Robust-DefReg and DefTransNet at every deformation level. The gains become particularly pronounced at larger deformations, where distance-only supervision is more likely to produce locally inconsistent displacement fields. For Robust-DefReg, the mean CD at deformation level 0.7 decreases from $39.49$ to $13.03$. For DefTransNet, the corresponding mean CD decreases from $33.57$ to $10.57$. Across all DeformedTissue deformation levels, DINE-PCA reduces the mean CD by approximately 27--69\% relative to the corresponding Stage-1 backbone, with the largest observed gains at moderate-to-large deformation. Within the DefTransNet family, DINE-PCA obtains the lowest mean CD at all levels, suggesting that the proposed prior can be beneficial when combined with a stronger registration backbone.

\begin{table}[t]
\centering
\caption{Registration accuracy on DeformedTissue under increasing deformation. Values are CD $\times 10^{-4}$, reported as mean $\pm$ standard deviation. Bold indicates the lowest mean CD within each backbone family. For DefTransNet, both DINE-PCA and DINE-Flow are compared with the same Stage-1 backbone. Lower is better. R-DefReg denotes Robust-DefReg.}
\label{tab:deftissue_deformation}
\scriptsize
\setlength{\tabcolsep}{2.8pt}
\renewcommand{\arraystretch}{1.15}
\resizebox{\textwidth}{!}{%
\begin{tabular}{l|cc||ccc}
\toprule
Level &
\makecell{Robust-DefReg\\\cite{monji2024robust}} &
\makecell{DINE-PCA\\(Robust-DefReg)} &
\makecell{DefTransNet\\\cite{monji2025deftransnet}} &
\makecell{DINE-PCA\\(DefTransNet)} &
\makecell{DINE-Flow\\(DefTransNet)} \\
\midrule
0.1 & 5.18 $\pm$ 3.03 & \textbf{3.78 $\pm$ 3.04} & 4.55 $\pm$ 3.17 & \textbf{3.12 $\pm$ 3.01} & 4.33 $\pm$ 3.87 \\
\hline
0.2 & 5.97 $\pm$ 3.20 & \textbf{4.10 $\pm$ 3.09} & 5.42 $\pm$ 3.14 & \textbf{3.39 $\pm$ 2.56} & 5.03 $\pm$ 3.51 \\
\hline
0.3 & 8.01 $\pm$ 7.69 & \textbf{4.46 $\pm$ 4.43} & 7.57 $\pm$ 6.32 & \textbf{4.24 $\pm$ 3.68} & 5.97 $\pm$ 4.55 \\
\hline
0.4 & 21.76 $\pm$ 11.62 & \textbf{8.67 $\pm$ 4.17} & 18.79 $\pm$ 10.70 & \textbf{7.54 $\pm$ 3.82} & 9.77 $\pm$ 4.52 \\
\hline
0.5 & 28.00 $\pm$ 9.39 & \textbf{10.47 $\pm$ 3.60} & 23.82 $\pm$ 8.32 & \textbf{8.74 $\pm$ 2.99} & 11.16 $\pm$ 3.64 \\
\hline
0.6 & 32.26 $\pm$ 7.98 & \textbf{9.92 $\pm$ 2.59} & 25.37 $\pm$ 6.42 & \textbf{8.18 $\pm$ 2.47} & 10.64 $\pm$ 2.95 \\
\hline
0.7 & 39.49 $\pm$ 6.61 & \textbf{13.03 $\pm$ 5.72} & 33.57 $\pm$ 7.41 & \textbf{10.57 $\pm$ 3.89} & 13.13 $\pm$ 4.57 \\
\hline
\end{tabular}%
}
\end{table}

On SynBench, DINE-PCA also obtains lower mean CD than the corresponding Stage-1 model for both backbones. For Robust-DefReg, this reduction is observed across all deformation levels, including the largest deformation level, where the mean CD decreases from $18.03$ to $3.72$. For DefTransNet, DINE-PCA gives lower mean CD than the Stage 1 baseline across the full deformation range. The gains are modest at the smallest deformation levels but reach approximately 59--79\% from moderate to severe deformation relative to the corresponding Stage-1 backbone. Within the DefTransNet family, DINE-Flow is competitive and achieves the lowest mean CD in selected settings, including $\sigma=0.5$ and $\sigma=0.8$. However, its mean CD values are less uniform across the full deformation range, especially at $\sigma=0.6$ and $\sigma=0.7$. Overall, the PCA-based full-field prior shows more uniform reductions across deformation levels, while the flow-based prior can be effective in specific deformation regimes.

\begin{table}[t]
\centering
\caption{Registration accuracy on SynBench under increasing deformation. Values are CD $\times 10^{-4}$, reported as mean $\pm$ standard deviation. Bold indicates the lowest mean CD within each backbone family. For DefTransNet, both DINE-PCA and DINE-Flow are compared with the same Stage-1 backbone. Lower is better. R-DefReg denotes Robust-DefReg.}
\label{tab:synbench_deformation}
\scriptsize
\setlength{\tabcolsep}{2.8pt}
\renewcommand{\arraystretch}{1.15}
\resizebox{\textwidth}{!}{%
\begin{tabular}{l|cc||ccc}
\toprule
Level &
\makecell{Robust-DefReg\\\cite{monji2024robust}} &
\makecell{DINE-PCA\\(Robust-DefReg)} &
\makecell{DefTransNet\\\cite{monji2025deftransnet}} &
\makecell{DINE-PCA\\(DefTransNet)} &
\makecell{DINE-Flow\\(DefTransNet)} \\
\midrule
0.1 & 1.36 $\pm$ 1.77 & \textbf{1.20 $\pm$ 1.74} & 1.18 $\pm$ 1.72 & \textbf{1.08 $\pm$ 1.57} & 1.85 $\pm$ 2.28 \\
\hline
0.2 & 1.86 $\pm$ 2.05 & \textbf{1.22 $\pm$ 1.70} & 1.00 $\pm$ 1.57 & \textbf{0.76 $\pm$ 1.35} & 0.93 $\pm$ 1.62 \\
\hline
0.3 & 4.15 $\pm$ 3.36 & \textbf{1.30 $\pm$ 1.69} & 3.49 $\pm$ 2.61 & 1.35 $\pm$ 1.58 & 2.45 $\pm$ 2.44 \\
\hline
0.4 & 7.75 $\pm$ 6.59 & \textbf{2.45 $\pm$ 2.37} & 6.29 $\pm$ 5.56 & \textbf{2.58 $\pm$ 2.44} & \textbf{2.58 $\pm$ 0.95} \\
\hline
0.5 & 14.73 $\pm$ 6.10 & \textbf{4.19 $\pm$ 2.61} & 9.80 $\pm$ 6.27 & 3.01 $\pm$ 2.52 & \textbf{1.01 $\pm$ 1.58} \\
\hline
0.6 & 16.15 $\pm$ 7.33 & \textbf{3.84 $\pm$ 3.08} & 12.94 $\pm$ 5.43 & \textbf{2.95 $\pm$ 2.14} & 8.30 $\pm$ 7.24 \\
\hline
0.7 & 18.05 $\pm$ 7.84 & \textbf{3.86 $\pm$ 3.17} & 14.07 $\pm$ 7.15 & \textbf{3.34 $\pm$ 2.82} & 9.73 $\pm$ 6.25 \\
\hline
0.8 & 18.03 $\pm$ 7.16 & \textbf{3.72 $\pm$ 2.95} & 15.17 $\pm$ 5.96 & 3.72 $\pm$ 2.57 & \textbf{2.12 $\pm$ 2.04} \\
\hline
\end{tabular}%
}
\end{table}

\begin{table}[t]
\centering
\caption{Robustness on DeformedTissue under outlier and Gaussian-noise corruptions. Values are CD $\times 10^{-4}$, reported as mean $\pm$ standard deviation. Bold indicates the lowest mean CD within each backbone family. Lower is better. R-DefReg denotes Robust-DefReg.}
\label{tab:corruption}
\scriptsize
\setlength{\tabcolsep}{2.8pt}
\renewcommand{\arraystretch}{1.15}
\resizebox{\textwidth}{!}{%
\begin{tabular}{l|l|cc||cc}
\toprule
Setting & Level &
\makecell{Robust-DefReg\\\cite{monji2024robust}} &
\makecell{DINE-PCA\\(Robust-DefReg)} &
\makecell{DefTransNet\\\cite{monji2025deftransnet}} &
\makecell{DINE-PCA\\(DefTransNet)} \\
\midrule
Outliers & 5\%  & 22.0 $\pm$ 10.0 & \textbf{12.0 $\pm$ 5.0} & 20.0 $\pm$ 8.0 & \textbf{11.0 $\pm$ 5.0} \\
\hline
Outliers & 25\% & 26.0 $\pm$ 8.0  & \textbf{11.0 $\pm$ 5.0} & 23.0 $\pm$ 6.0 & \textbf{11.0 $\pm$ 4.0} \\
\hline
Outliers & 45\% & 29.0 $\pm$ 6.0  & \textbf{10.0 $\pm$ 3.0} & 26.0 $\pm$ 6.0 & \textbf{11.0 $\pm$ 3.0} \\
\hline
Noise & 0.01 & 24.0 $\pm$ 5.0 & \textbf{8.0 $\pm$ 3.0} & 20.0 $\pm$ 4.0 & \textbf{7.0 $\pm$ 2.0} \\
\hline
Noise & 0.03 & 33.0 $\pm$ 3.0 & \textbf{6.0 $\pm$ 2.0} & 29.0 $\pm$ 2.0 & \textbf{6.0 $\pm$ 1.0} \\
\hline
Noise & 0.04 & 30.0 $\pm$ 2.0 & \textbf{6.0 $\pm$ 2.0} & 29.0 $\pm$ 2.0 & \textbf{5.0 $\pm$ 1.0} \\
\hline
\end{tabular}%
}
\end{table}

\textbf{Robustness to outliers and noise.}
Table~\ref{tab:corruption} reports robustness on DeformedTissue under outlier contamination and Gaussian noise. DINE-PCA obtains lower mean CD than the corresponding Stage-1 model for both Robust-DefReg and DefTransNet across all corruption settings. Under outlier contamination, the DINE-regularized models maintain lower mean CD as the outlier ratio increases from 5\% to 45\%. With 45\% outliers, Robust-DefReg obtains a mean CD of $29.0$, while DINE-PCA with Robust-DefReg obtains $10.0$. Similarly, DefTransNet obtains $26.0$, whereas DINE-PCA with DefTransNet obtains $11.0$.

The same trend is observed under Gaussian noise. At noise level 0.03, the mean CD decreases from $33.0$ to $6.0$ for Robust-DefReg and from $29.0$ to $6.0$ for DefTransNet. At noise level 0.04, DINE-PCA with DefTransNet obtains the lowest mean CD of $5.0$. These results suggest that the proposed prior can improve not only registration under clean deformation, but also robustness to common acquisition artifacts. This is particularly important for biomedical 3D scanning, where point clouds often contain noisy surfaces, outliers, and local reconstruction errors.

\begin{figure}[ht]
\centering
\includegraphics[width=0.85\textwidth]{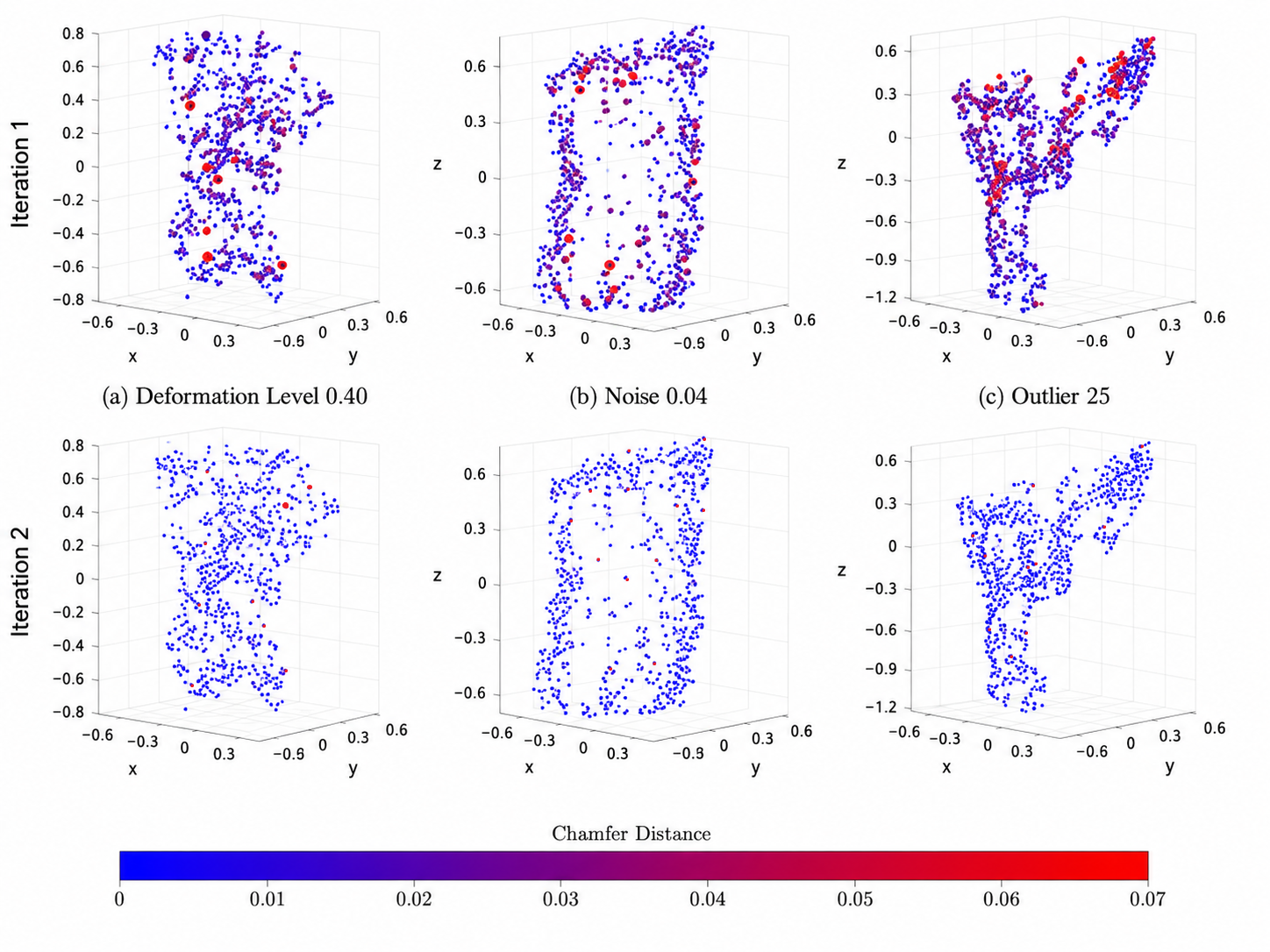}
\caption{Qualitative comparison of DefTransNet registration before and after PCA-prior refinement. Points are colored by Chamfer distance to the target, from blue for low error to red for high error. Stage 2 reduces scattered high-error regions across deformation, noise, and outlier settings.}
\label{fig:qualitative}
\end{figure}

\textbf{Qualitative and distributional analysis.}
Figure~\ref{fig:qualitative} visualizes representative DefTransNet results before and after PCA-prior refinement. The Stage 2 model reduces scattered high-error regions and produces more uniform alignment under deformation, noise, and outlier contamination. This qualitative behavior is consistent with the quantitative results in Tables~\ref{tab:deftissue_deformation} and~\ref{tab:corruption}, where DINE-PCA obtains lower mean CD on DeformedTissue.

\begin{figure}[ht]
\centering
\includegraphics[width=0.8\textwidth]{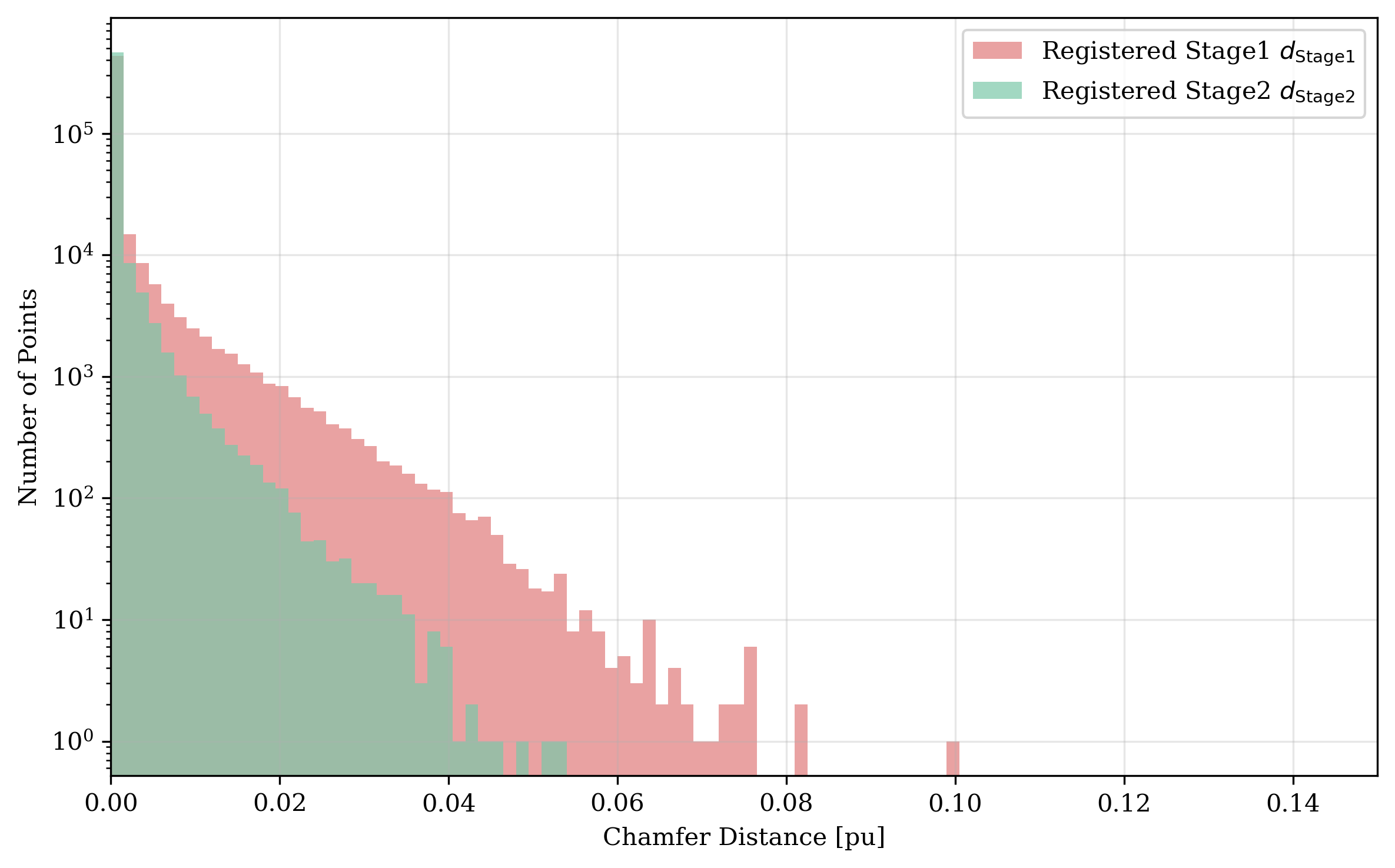}
\caption{Distribution of per-point Chamfer distances on SynBench for DefTransNet. The PCA prior suppresses the high-error tail observed in the distance-only Stage 1 model.}
\label{fig:histogram}
\end{figure}

Figure~\ref{fig:histogram} shows the per-point Chamfer distance distribution on SynBench for DefTransNet. Stage 1 has a long tail of high residuals, while Stage 2 with the PCA prior suppresses this tail. This indicates that the prior does more than reduce average CD: it reduces the frequency of severe local registration failures. Such behavior is important for medical shape registration, where localized misalignment can be relevant even when the average CD is low.

\section{Discussion}

The results support the main hypothesis of this paper: local distance supervision alone is not sufficient for robust soft-tissue point cloud registration under large deformation. On both DeformedTissue and SynBench, the mean CD of distance-only backbones increases as deformation becomes larger. Chamfer distance measures local proximity, but it does not encode whether the full DVF is plausible. Adding a learned deformation prior restricts the solution space toward deformation fields that are more likely under the learned prior and therefore can reduce the tendency to fit noisy or ambiguous local correspondences.

The comparison between PCA and flow priors highlights the importance of spatial correlation. The flow prior can model a flexible distribution over individual displacement vectors, but it treats these vectors independently. Therefore, it cannot distinguish between a coherent tissue motion pattern and a spatially inconsistent collection of locally plausible displacements. In contrast, the PCA prior operates on the full vectorized DVF and captures correlations between point displacements across the entire surface. This may explain why DINE-PCA shows more uniform reductions across deformation levels and corruptions in our experiments.

For medical shape analysis, this result is important because soft-tissue deformation is not only a point-wise correspondence problem. It is a structured shape transformation with global dependencies. The DeformedTissue experiments demonstrate this point in a real soft-tissue setting, while SynBench provides controlled evidence that the effect is not limited to one dataset. The proposed framework is also architecture-agnostic: the same prior strategy gives lower mean CD for both Robust-DefReg and DefTransNet. This suggests that statistical deformation modeling can be added as a lightweight and interpretable regularization layer on top of modern geometric learning systems.

The qualitative and residual analyses provide additional insight. The PCA prior reduces scattered high-error regions and suppresses the high-error tail of the per-point Chamfer distribution. This suggests that the prior does not simply lower the mean CD by small local adjustments; it can reduce severe local failures that are typical under noise, outliers, and large deformation. These findings also indicate that future improvements may come from combining DINE with robust or non-quadratic Chamfer variants, which could further improve robustness to outliers, density changes, and heavy-tailed residuals.

\textbf{Limitations.}
The PCA prior assumes that the dominant deformation manifold is approximately linear in the selected latent space. Highly nonlinear or multi-modal deformations may require more expressive full-field priors, such as a VAE or a normalizing flow over the complete DVF. The prior is learned from Stage-1 predictions rather than ground-truth deformation fields; if the first-stage model is systematically biased, the learned prior may inherit this bias. In this work, the Stage-1 fields are used only as an empirical estimate of dominant recurring deformation modes, and the prior acts as a regularizer rather than a replacement for the geometric data term. The Gaussian PCA prior should be interpreted as a modeling approximation rather than as a verified distributional property of the deformation fields. In addition, we report mean and standard deviation values without formal paired significance testing, so the observed differences should be interpreted as empirical performance trends.

\section{Conclusion}

We presented DINE, a MAP-based framework for learning statistical deformation priors in non-rigid soft-tissue point cloud registration. The method uses high-performing registration backbones and augments their distance-based objective with a learned negative log-prior over DVFs. The full-field PCA Gaussian prior provides stable empirical gains because it captures spatial correlations across the complete deformation field, while the per-vector flow prior is less uniform across settings due to its lack of inter-point structure. Experiments on DeformedTissue and SynBench show lower mean CD under large deformation, noise, and outliers. These findings suggest that statistical deformation modeling remains relevant in modern geometric deep learning and can improve the reliability of shape registration for deformable medical structures.

\bibliographystyle{splncs04}
\bibliography{references}
\end{document}